\documentclass[11pt]{article}
\usepackage{acl}
\usepackage{times}
\usepackage{latexsym}
\usepackage[T1]{fontenc}
\usepackage[utf8]{inputenc}
\usepackage{microtype}
\usepackage{graphicx}
\usepackage{booktabs}
\usepackage{amsmath}
\usepackage{url}
\usepackage{xcolor}
\usepackage{multirow}
\usepackage{tikz}
\usetikzlibrary{positioning,arrows.meta}
\tikzset{ic/.style={baseline=-0.28cm}}
\newcommand{\stagecontent}[3]{%
  \begin{minipage}[c]{2.35cm}\centering
    \setlength{\parskip}{0pt}\setlength{\parindent}{0pt}%
    \begin{minipage}[t][1.0cm]{2.35cm}\centering\vspace*{0pt}#1\end{minipage}\par
    \begin{minipage}[c][1.0cm]{2.35cm}\centering\textbf{#2}\end{minipage}\par
    \begin{minipage}[b][1.0cm]{2.35cm}\centering{\scriptsize\itshape #3}\end{minipage}%
  \end{minipage}%
}
\newcommand{\menokewall}{53 min}
\newcommand{\menokegpu}{0.9}
\newcommand{\menokeco}{0.14}
\newcommand{\valmenovalj}{0.438}
\newcommand{\valmenovala}{0.456}
\newcommand{\valmenovale}{0.899}
\newcommand{\qsevenba}{0.735}
\newcommand{\qsevenbe}{0.956}
\newcommand{\qsevenbj}{0.691}
\newcommand{\qsevenpota}{0.757}
\newcommand{\qsevenpote}{0.646}
\newcommand{\qsevenpotj}{0.497}
\newcommand{\menozeroonea}{0.420}
\newcommand{\menozeroonee}{0.895}
\newcommand{\menozeroonej}{0.359}
\newcommand{\menozeroonepota}{0.674}
\newcommand{\menozeroonepote}{0.934}
\newcommand{\menozeroonepotj}{0.641}

\newif\iffinalversion
\finalversiontrue

\iffinalversion
\title{When Harness Beats Scale, and When Reading Beats Both}
\author{
  \textbf{Ivan Bondarenko}$^{1}$ \quad
  \textbf{Nikolay O. Nikitin}$^{2}$ \\[4pt]
  $^{1}$Novosibirsk State University \quad
  $^{2}$ITMO University \\[2pt]
  \texttt{i.bondarenko@g.nsu.ru} \quad \texttt{nicl.nno@gmail.com}
}
\else
\title{When Harness Beats Scale, and When Reading Beats Both}
\author{Anonymous}
\fi

\begin{document}
\maketitle

\begin{abstract}
We describe our system for DocSem, the document-grounded quantitative reasoning shared task at DocInsights 2026, and analyze why it succeeded on labeled data and failed on the test set.
The pipeline pairs hybrid block retrieval with Program-of-Thoughts (PoT) generation executed in a sandboxed interpreter, self-consistency sampling, and entity enrichment from chunk-level knowledge graphs.
On our held-out split, application architecture moved the metrics far more than model scale did: PoT added 0.282 joint accuracy to a compact 7B model but at most 0.005 to a 72B model, and a 27B model with the full harness matched the 72B (0.884 vs.\ 0.873) at roughly 2.7$\times$ fewer parameters and a quarter of the CO$_2$.
We read this through a distinction between world knowledge, which scales steeply with parameters, and language knowledge, which scales gently, and show that structured-output training makes a compact model harness-ready rather than merely small.
On the raster, watermarked test PDFs the same system collapsed to 13.58\% joint (rank 149 of 163); a controlled re-rendering of the validation set reproduces the OCR half of the collapse while bounding what the simulation misses.
Auditing the physical nature of evaluation inputs precedes architecture, and the leaderboard's bimodality is consistent with reading quality, not reasoning, having separated the field.
\end{abstract}

\section{Introduction}
\label{sec:intro}

DocSem asks for two things at once: given a PDF and a paraphrased query, compute a numeric answer strictly from one embedded quantitative passage, and name the exact layout block identifiers that justify it.
Scoring is strict on both axes, and the primary ranking metric, joint exact accuracy, rewards systems that find the passage and reproduce its arithmetic without contamination from the dozens of unrelated numbers that surround it.

Our system builds on RAGU, a graph retrieval-augmented generation toolkit, and entered the test phase at 0.901 joint exact accuracy on our internal held-out split and 0.853 on the organizers' validation portal.
The test phase returned 13.58\%, rank 149 of 163 teams.
This paper documents both halves of that gap, because each half carries a transferable lesson.

On labeled data, application architecture dominated model scale.
Switching a compact 7B model from direct answering to Program-of-Thoughts (PoT), where the model writes a program that a sandboxed interpreter executes \citep{chen2023program}, raised its joint accuracy by 0.282, while the same switch moved a 72B model by at most 0.005; a 27B model with PoT then matched the 72B answering directly (0.884 vs.\ 0.873 joint).
A lineage comparison against the base Qwen2.5-7B-Instruct \citep{yang2024qwen2} qualifies the lever: the same harness that rescues the domain-adapted 7B model taxes the base model's citation discipline, so PoT is model-dependent; the divergence is portable, reappearing in every rigid output format we tested (JSON validity, answer markers, evidence identifiers).
Graph artifacts helped only in distilled form, and only at scale: appending the entities nearest to the query, out of chunk-level graphs built by a 7B extractor, added 0.017 joint on the 27B line and 0.012 on the 72B, but the same enrichment hurt the compact 7B line, dropping the base from 0.497 to 0.448 joint and the adapted model from 0.641 to 0.011, while fuller graph views subtracted up to 0.011.

We read this asymmetry through a distinction between two kinds of competence an LLM can supply: knowledge about the world, which scales steeply with parameter count, and knowledge about language---comprehension, extraction, faithful reformatting---which scales far more gently.
A harness that routes world knowledge to external tools (an interpreter for arithmetic, a retriever for passage search, an ontology for fact structure) leaves the model only the linguistic residue, and it is on that residue that a compact model is competitive: our 7B line trails the 72B baseline by 0.10 on instruction following and 0.05 on flexible arithmetic, gaps far smaller than any world-knowledge comparison would show.
The same harness reaches 0.64 joint on a single GPU at 0.14\,kgCO$_2$e for its check runs, against eight GPUs and 0.53\,kgCO$_2$e for the 72B line (Appendix~\ref{app:cost}).

On the test set, none of this mattered, because the test PDFs are raster scans with watermarks, unlike the born-digital training and validation documents, and our reading pipeline degraded their text beyond what retrieval and arithmetic could recover.
Three facts pin down the diagnosis.
Answers flipped on 89.6\% of tasks when we replaced OCR with page-level vision transcription: the flip measures the input, with the reasoning stack untouched.
Evidence sets, singleton in all training labels, grew to three or more blocks on 32.5\% of test tasks, a signature of a model that cannot locate the passage and hedges.
And the final leaderboard is bimodal, with a 22-team spike at 67.46\% and a long reading-failure tail in which we sit.

Our contributions:
\begin{itemize}\itemsep2pt
\item a DocSem system in which architectural choices (PoT execution, self-consistency, distilled graph context) moved the metrics further than any model-scale increase we could afford;
\item a 7B lineage comparison (Meno-Lite-0.1 against its base Qwen2.5-7B-Instruct) showing the PoT lever is model-dependent: it rescued the domain-adapted model and taxed the base model's citation discipline;
\item a cross-task reading of that lineage (instruction following, arithmetic, world knowledge, JSON generation, extraction) showing that structured-output training confers a language-independent discipline in rigid formats, while the base model keeps an edge in free-form instruction;
\item retrieval and context ablations showing that only the top-$k$ distillation of chunk-level graphs helps, while fuller graph views hurt;
\item a submission-level post-mortem of three test attempts, plus a controlled degradation study that reproduces the OCR half of the test collapse on validation inputs and bounds what the simulation misses.
\end{itemize}

\section{Task and the Data Shift We Missed}
\label{sec:data}

Each task provides a PDF and a user query paraphrasing an embedded GSM-style quantitative passage \citep{singh2026gsmsem,cobbe2021gsm8k}. Documents contain a background narrative, tables, dates, and numeric facts; roughly one block in ten holds tabular content.
Blocks carry visible identifiers (a token ending in a colon); a submission must return the answer and the set of justifying identifiers, matched exactly against the gold set.
In all 908 training labels the gold evidence is a single block and the answer is numeric, which fixed two design choices: output exactly one identifier unless the passage demands more, and emit canonical decimal strings.

\begin{figure*}[t]
\centering
\begin{minipage}[t]{0.49\textwidth}
\centering
\includegraphics[width=\linewidth,height=0.62\textheight,keepaspectratio]{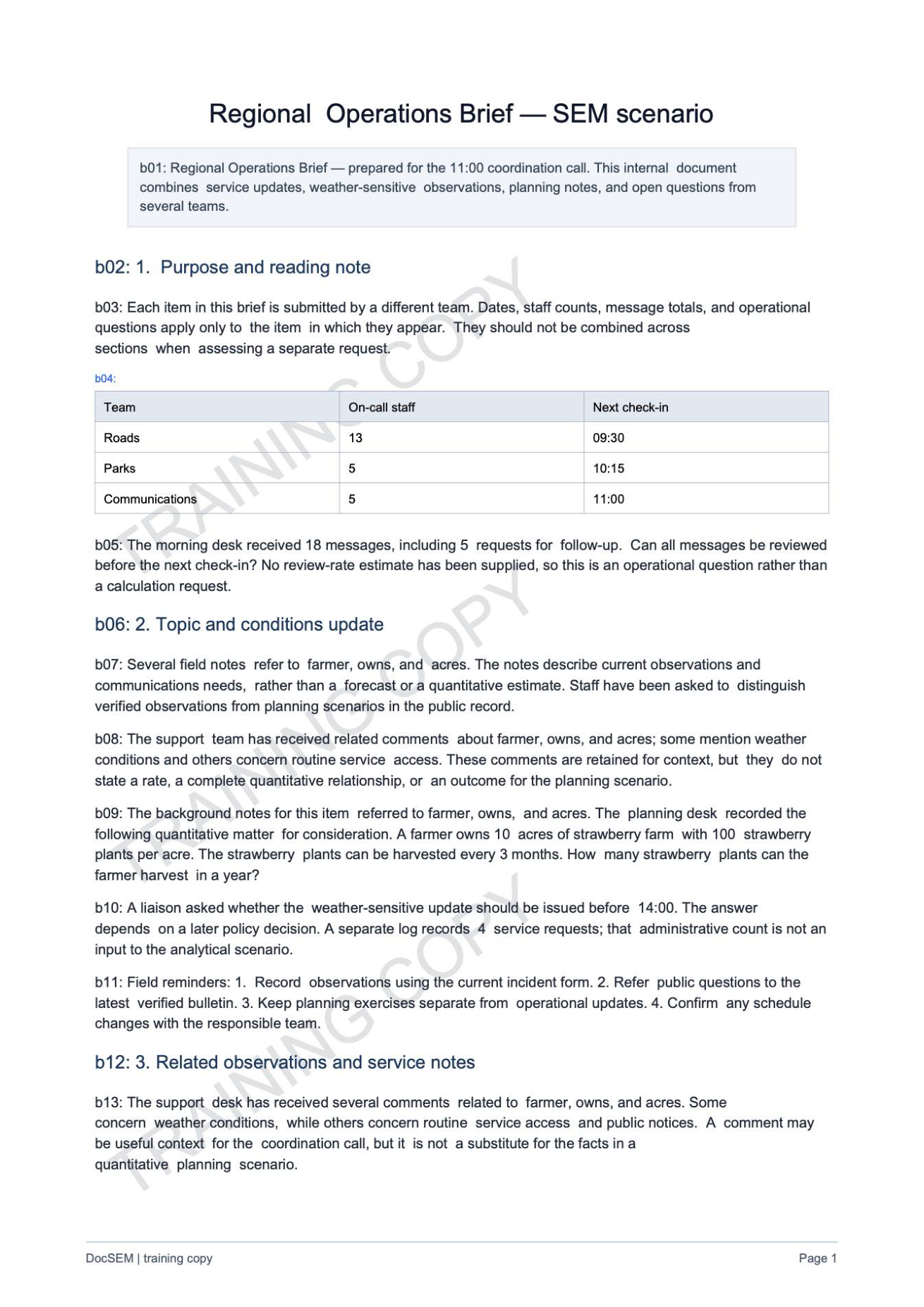}\\[2pt]
{\footnotesize\textbf{(a) validation, clean rendering}}\\[4pt]
\begin{minipage}{0.98\linewidth}\scriptsize
\texttt{b01: Regional Operations Brief --- prepared for the 11:00 coordination call.}\\[1pt]
\texttt{b05: The morning desk received 18 messages, including 5 requests for follow-up.}\\[1pt]
\texttt{b06: 2. Topic and conditions update}
\end{minipage}
\end{minipage}
\hfill
\begin{minipage}[t]{0.49\textwidth}
\centering
\includegraphics[width=\linewidth,height=0.62\textheight,keepaspectratio]{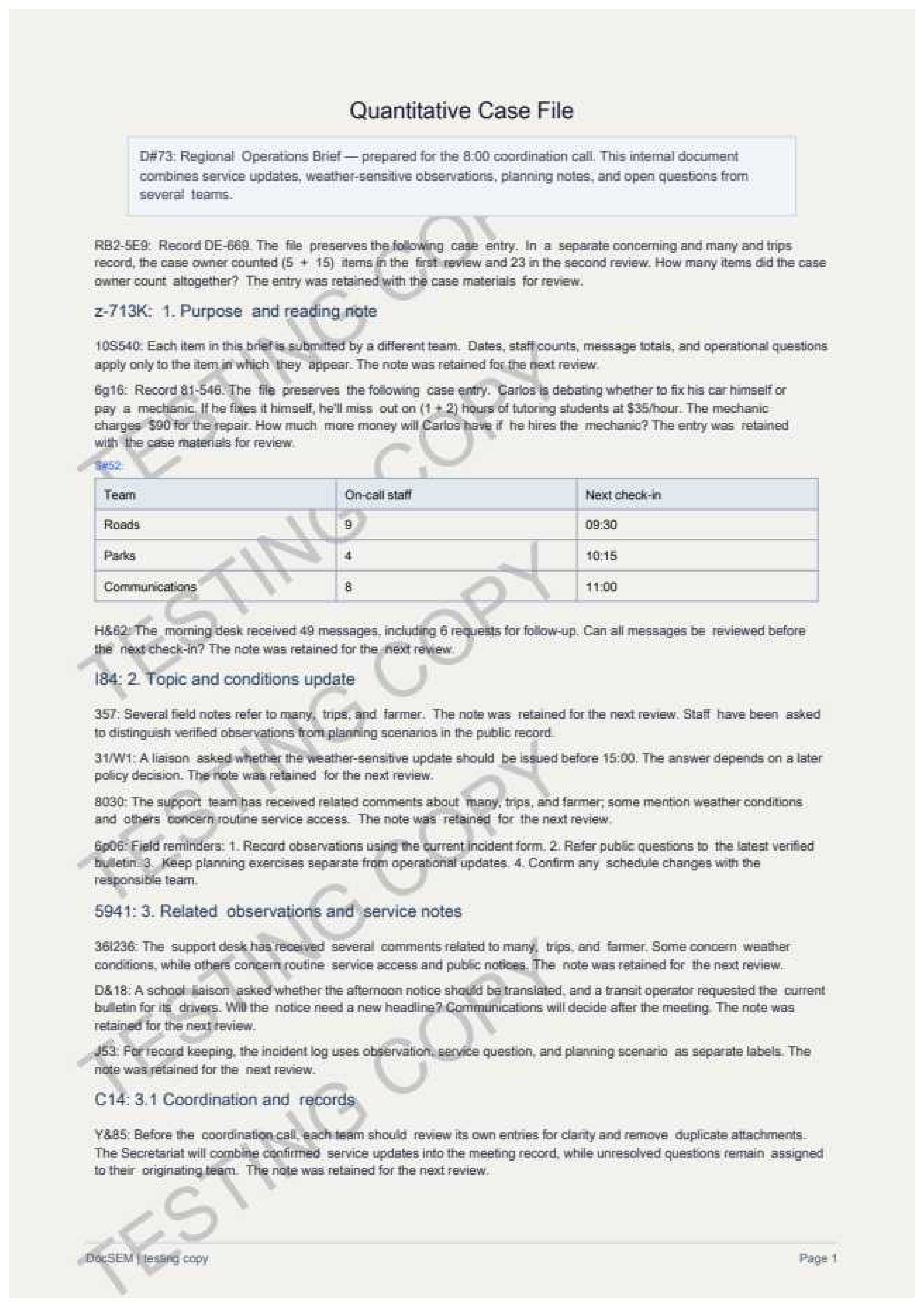}\\[2pt]
{\footnotesize\textbf{(b) test, raster scan with watermark}}\\[4pt]
\begin{minipage}{0.98\linewidth}\scriptsize
\texttt{\#72: Regional Operations Brief --- prepared for the 8:00 coordination call.}\\[1pt]
\texttt{H\&62$^\circ$ The received 49 messages, incl. 6 requests for follow-up.}\\[1pt]
\texttt{2-713K: 1. Purpose and reading note}
\end{minipage}
\end{minipage}
\caption{The same task under two input regimes. (a) A validation page renders cleanly: block identifiers survive and Tesseract recovers the text. (b) A test page is a low-resolution raster scan stamped with a diagonal ``TESTING COPY'' watermark and running headers; Tesseract garbles the identifiers (\texttt{\#72:}, \texttt{H\&62$^\circ$}, \texttt{2-713K:}) so evidence can no longer match gold labels.}
\label{fig:reading}
\end{figure*}

Our exploratory analysis of the training data verified the things we thought to check: no duplicate documents across splits (908/217/1730 unique PDFs), 99.78\% recoverability of gold blocks by our parser, a block-length tail short enough (p99 = 638 tokens) to avoid truncation in an 8k context.
The test snapshot differed in one more way, which we measured late: its PDFs are low-resolution raster scans, one image per page, stamped with a diagonal ``TESTING COPY'' watermark and running headers, while training and validation pages are clean renderings of born-digital documents.
Tesseract \citep{tesseract} on these pages yields a median service-word share of 0.091 versus 0.224 on validation (Figure~\ref{fig:reading}), and our later audit found pages where the vision model transcribed the watermark in loops until it hit the token ceiling.
We verified checksums before the deadline; we did not look at the pixels.
What we now audit before designing the reading stack: born-digital versus raster pages, the presence of a text layer, image resolution, watermarks and running headers, and the distribution of page and block counts---a check that would have caught the test shift in minutes.

\section{System}
\label{sec:system}

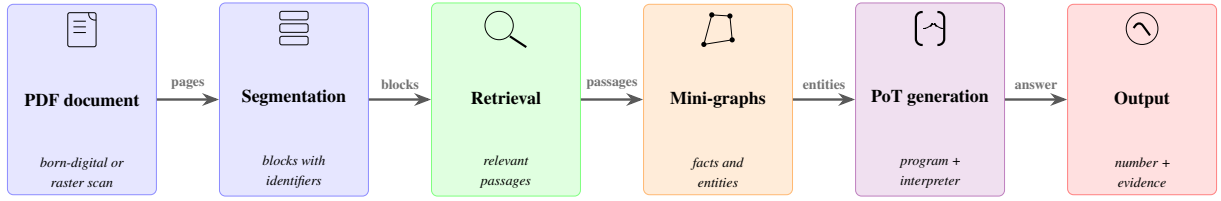
\begin{figure*}[t]
\centering
\resizebox{\textwidth}{!}{%
\begin{tikzpicture}[
    box/.style={draw, rounded corners=3pt, inner sep=4pt, font=\footnotesize},
    arr/.style={-{Stealth[length=3mm]}, very thick, black!65},
    lbl/.style={font=\scriptsize\bfseries, text=black!55},
]
\node[box, fill=blue!10, draw=blue!45] (pdf) {\stagecontent{%
  \tikz[ic]{\draw[rounded corners=1pt] (0,0) rectangle (0.5,0.6); \draw (0.37,0.6)--(0.37,0.47)--(0.5,0.47); \draw (0.1,0.45)--(0.32,0.45); \draw (0.1,0.35)--(0.32,0.35); \draw (0.1,0.25)--(0.32,0.25);}%
}{PDF document}{born-digital or\\raster scan}};
\node[box, fill=blue!10, draw=blue!45, right=1.1cm of pdf] (blocks) {\stagecontent{%
  \tikz[ic]{\foreach \y in {0,0.22,0.44} {\draw[rounded corners=1pt] (0,\y) rectangle (0.5,\y+0.15);}}%
}{Segmentation}{blocks with\\identifiers}};
\node[box, fill=green!12, draw=green!50, right=1.1cm of blocks] (retr) {\stagecontent{%
  \tikz[ic]{\draw (0.02,0.38) circle (0.25); \draw[line width=1.1pt] (0.21,0.19)--(0.5,0);}%
}{Retrieval}{relevant\\passages}};
\node[box, fill=orange!15, draw=orange!55, right=1.1cm of retr] (graph) {\stagecontent{%
  \tikz[ic]{\fill (0,0) circle (1.3pt); \fill (0.5,0.08) circle (1.3pt); \fill (0.14,0.52) circle (1.3pt); \fill (0.46,0.5) circle (1.3pt); \draw (0,0)--(0.5,0.08); \draw (0,0)--(0.14,0.52); \draw (0.14,0.52)--(0.46,0.5); \draw (0.5,0.08)--(0.46,0.5);}%
}{Mini-graphs}{facts and\\entities}};
\node[box, fill=violet!12, draw=violet!50, right=1.1cm of graph] (pot) {\stagecontent{%
  \tikz[ic]{\draw[line width=1.1pt] (0.12,0)--(0,0)--(0,0.58)--(0.12,0.58); \draw[line width=1.1pt] (0.38,0)--(0.5,0)--(0.5,0.58)--(0.38,0.58); \draw (0.17,0.29)--(0.24,0.29)--(0.31,0.42)--(0.38,0.29)--(0.44,0.29);}%
}{PoT generation}{program +\\interpreter}};
\node[box, fill=red!12, draw=red!50, right=1.1cm of pot] (out) {\stagecontent{%
  \tikz[ic]{\draw (0.28,0.3) circle (0.28); \draw[line width=1.1pt] (0.13,0.31)--(0.23,0.44)--(0.42,0.18);}%
}{Output}{number +\\evidence}};
\draw[arr] (pdf) -- (blocks) node[midway, above=1pt, lbl] {pages};
\draw[arr] (blocks) -- (retr) node[midway, above=1pt, lbl] {blocks};
\draw[arr] (retr) -- (graph) node[midway, above=1pt, lbl] {passages};
\draw[arr] (graph) -- (pot) node[midway, above=1pt, lbl] {entities};
\draw[arr] (pot) -- (out) node[midway, above=1pt, lbl] {answer};
\end{tikzpicture}}
\caption{The DocSem pipeline. Raster test pages are transcribed by a VLM before block segmentation; a hybrid retriever selects relevant passages; RAGU builds chunk-level graphs whose entities enrich the prompt; Program-of-Thoughts returns the answer and evidence.}
\label{fig:pipeline}
\end{figure*}

\subsection{Document ingestion}
Figure~\ref{fig:pipeline} sketches the pipeline.
Blocks are segmented by the generic rule ``identifier token up to the first colon'', which copies opaque test tokens verbatim; tables are linearized row-wise as \texttt{column: value} lines inside their block.
Training and validation PDFs are parsed directly; raster test pages were transcribed page-by-page by Qwen2.5-VL-7B \citep{bai2025qwen25vl} at 2$\times$ zoom, with pages distributed across service instances and stitched by page index.
A targeted repair pass with Qwen3.8-27B \citep{qwen38} re-read 38 damaged pages (26 documents): three empty ones and 35 degenerate repetition loops on near-empty watermarked pages.

\subsection{Retrieval}
Each deduplicated document gets its own index; a block is a chunk.
Dense vectors come from gte-multilingual-base \citep{zhang-etal-2024-mgte}, sparse from BM42, merged by reciprocal rank fusion \citep{cormack2009rrf} in Qdrant; a gte-multilingual-reranker cross-encoder cuts the fused top-8 to the 3 blocks shown to the generator.
The pair was chosen by an evidence-recall benchmark on our held-out split: bge-m3 reached recall@1 of 0.348, gte 0.901; after reranking, gte saturates at 1.000 and bge reaches 0.807 (Appendix~\ref{app:retrieval}).

\subsection{Program-of-Thoughts generation}
The generator writes a Python program that reads only the displayed blocks, and a sandboxed interpreter executes it; the final variable becomes the answer, and the cited identifiers become evidence.
Structured output (a JSON schema over reasoning, program, and evidence) is enforced server-side by vLLM \citep{kwon2023efficient}.
Self-consistency \citep{wang2023selfconsistency} samples $k{=}5$ programs at $T{=}0.8$ and votes on the executed answers; identifiers snap to valid block tokens when transcription noise corrupts them.

\subsection{Graph context and final assembly}
Chunk-level mini-graphs are extracted from the retrieved passages by Meno-Lite-0.2, a compact domain-adapted Qwen2-family model and the successor of the released Meno-Lite-0.1,\footnote{\url{https://huggingface.co/bond005/meno-lite-0.1}; 0.2 is being prepared for release.} with a numeric ontology (NEREL entity types \citep{loukachevitch2021nerel} extended with QUANTITY and RATE).
The engine underneath, from block indexing to search, is RAGU \citep{komarov2026ragu}.\footnote{\url{https://github.com/RaguTeam/RAGU}}
The generator prompt is enriched with the entities nearest to the query in embedding space.
The final test submission merged three votes by majority (the base PoT run, the key-entity run, and a multimodal escalation that shows page images of non-unanimous tasks to Qwen3.8-27B).
While test scores were still hidden, an external GLM-5.3-Flash judge probed 60 test tasks; the released leaderboard later superseded these estimates.

\section{Results on Labeled Data}
\label{sec:results}

\begin{table}[t]
\centering\footnotesize
\resizebox{\columnwidth}{!}{%
\begin{tabular}{lccc}
\toprule
Configuration (check, 181 tasks) & Ans. & Evid. EM & Joint \\
\midrule
Qwen2.5-7B-Instruct, direct & \qsevenba{} & \qsevenbe{} & \qsevenbj{} \\
Qwen2.5-7B-Instruct, PoT & \qsevenpota{} & \qsevenpote{} & \qsevenpotj{} \\
Meno-Lite-0.1 7B, direct & \menozeroonea{} & \menozeroonee{} & \menozeroonej{} \\
Meno-Lite-0.1 7B, PoT & \menozeroonepota{} & \menozeroonepote{} & \menozeroonepotj{} \\
Qwen2.5-72B, direct, $k{=}1$ & 0.867 & 1.000 & 0.867 \\
Qwen2.5-72B, direct, $k{=}3$ & 0.873 & 1.000 & 0.873 \\
\quad full document, $k{=}3$ & 0.884 & 1.000 & 0.884 \\
Qwen2.5-72B, PoT, $k{=}3$ & 0.873 & 1.000 & 0.873 \\
Qwen2.5-72B, PoT, $k{=}5$, $T{=}0.8$ & 0.878 & 1.000 & 0.878 \\
Qwen3.8-27B, PoT, $k{=}5$ & 0.884 & 1.000 & 0.884 \\
\quad + key entities (top-3) & 0.901 & 1.000 & \textbf{0.901} \\
\quad + mini-graph (12 ent.) & 0.878 & 1.000 & 0.878 \\
\quad + entities, edges & 0.890 & 1.000 & 0.890 \\
\bottomrule
\end{tabular}}
\caption{Main line on the internal check split. PoT moves a compact model further than any scale jump; graph context helps only in top-3 distilled form.}
\label{tab:main}
\end{table}

Table~\ref{tab:main} traces the generation line, and its reading is asymmetric by model size and by model origin.
Consider the 7B pair first, where the same two protocols run on Meno-Lite-0.1 and on its root ancestor Qwen2.5-7B-Instruct.
Direct answering splits them: the unmodified base reaches 0.691 joint (0.735 answers, 0.956 evidence exact match), while the domain-adapted model lands at 0.359, a drop consistent with its declared focus on Russian-language RAG and extraction skills.
Program-of-Thoughts then moves them in opposite directions: it lifts Meno-Lite by 0.282 joint (0.359 to 0.641, with evidence discipline rising to 0.934) and pulls the base down by 0.194 (0.691 to 0.497), because the base keeps computing (0.757 answers) but stops citing carefully (0.646 evidence exact match) once it writes programs.
The harness lever is real but model-dependent: it pays most where the model is weakest, and it can tax a strength.

At the top of the scale the same lever moves nothing: the 72B model scores 0.873 with direct answering and 0.873 with PoT at identical sampling ($k{=}3$, $T{=}0.7$), and 0.878 at the champion sampling ($k{=}5$, $T{=}0.8$).
PoT is an equalizer, and the 27B model with the full harness (0.884, or 0.901 with key entities) matches or passes every 72B row in the table at roughly 2.7$\times$ fewer parameters.
Self-consistency behaves the same way: $k{=}3$ over $k{=}1$ adds 0.006 on the 72B line, against 0.022 for $k{=}5$ over $k{=}3$ on the compact line.
As shown in Appendix~\ref{app:registry}, PoT demonstrations trade answer accuracy for perfect evidence.

The asymmetry has an explanation that guided our design.
DocSem documents are synthetic: cities, agencies, and numbers are invented for the benchmark, so world knowledge memorized in parameters buys nothing, and the residual demands on the LLM are linguistic (comprehension of a paraphrased query, selection of stated facts, faithful program writing), while arithmetic, passage search, and fact structuring are routed to the interpreter, the retriever, and the ontology.
This is the design hypothesis behind Meno-Lite, a 7B line trained to read rather than to memorize \citep{bondarenko2026menolite}.
DocSem, an English benchmark outside that model's primary domain, tests the hypothesis from its weak side: the harness recovers most of the distance the domain adaptation had cost (0.359 to 0.641 joint), while the same harness adds at most 0.005 to a model ten times larger.
Retrieval ablations and the full-document baseline complete the picture in the appendix.
The dense retriever does almost all the selection work: recall@3 is 0.9945 for dense alone and 0.0663 for BM42, because the paraphrase queries are built to avoid the target passage's wording; the cross-encoder closes the residual gap (1.000 at top-3).
The full-document row of Table~\ref{tab:main} adds an uncomfortable fact: with the 72B generator, feeding all 23--42 blocks matches curated top-3 selection (0.884 vs.\ 0.873, within noise).
Block selection earned its place on the compact line, where context discipline and cost bind, and it kept evidence attribution exact on every run; for a strong generator on short documents it was accuracy-neutral.

\begin{table}[t]
\centering\footnotesize
\resizebox{\columnwidth}{!}{%
\begin{tabular}{lccc}
\toprule
Validation portal (217 tasks) & Ans. & Evid. EM & Joint \\
\midrule
Meno-Lite-0.1 + PoT & \valmenovala{} & \valmenovale{} & \valmenovalj{} \\
Qwen2.5-7B-Instruct, direct & 0.659 & 0.959 & 0.636 \\
Meno-Lite-0.2 (unreleased) + PoT, track C (App.~\ref{app:retrieval}) & 0.682 & 0.949 & 0.641 \\
Qwen2.5-72B, prompt v1 & 0.774 & 1.000 & 0.774 \\
Qwen2.5-72B, prompt v2 & 0.825 & 1.000 & 0.825 \\
Qwen3.8-27B + PoT & 0.848 & 1.000 & 0.848 \\
\quad + permutation ensemble & -- & -- & 0.853 \\
\quad + key entities & -- & -- & 0.853 \\
\bottomrule
\end{tabular}}
\caption{Portal-scored validation submissions. The 27B PoT line overtakes the 72B direct line; graph enrichment adds 0.005 at portal precision.}
\label{tab:val}
\end{table}

The validation portal (Table~\ref{tab:val}) confirms the ordering on organizer-held labels: the 27B PoT configuration passes the 72B direct one, and the key-entity enrichment holds a small positive effect.
The compact line's own row carries a caution about off-domain generalization: Meno-Lite-0.1 with PoT drops from 0.641 joint on check to \valmenovalj{} on the portal, a 0.203 gap nearly four times the base model's 0.055 (0.691 to 0.636), and the loss is concentrated in answers (0.674 to \valmenovala{}) while evidence discipline barely moves (0.934 to \valmenovale{}).
Majority voting over four context variants (base, key entities, full mini-graph, entities-plus-edges) scored 0.890--0.901 joint on check against 0.901 for the best single vote, which is why the three-vote merge of the final submission stayed a hedge against transcription noise rather than an accuracy device.
Single-row differences on 181 tasks carry 95\% confidence intervals of roughly $\pm$4.5 points (0.901 maps to 0.849--0.937), so we read the table by its ordering across model lines and samplers rather than by any one delta.

\subsection{Structured-Output Training as a Harness Prerequisite}
\label{sec:lineage}

The 7B comparison is not confined to DocSem: we ran the base and the adapted model through instruction following (IFEval \citep{zhou2023ifeval}), arithmetic (GSM8K \citep{cobbe2021gsm8k}), world knowledge (MMLU \citep{hendrycks2021mmlu}, English and Russian), and structured generation (JSONSchemaBench \citep{geng2025jsonschemabench}) with lm-evaluation-harness \citep{gao2024lmharness} at temperature 0 and chat templates applied server-side.
Table~\ref{tab:lineage} condenses the outcome, and two patterns stand out.

\begin{table}[t]
\centering\footnotesize
\resizebox{\columnwidth}{!}{%
\begin{tabular}{lcc}
\toprule
Benchmark & Qwen2.5-7B-Instruct & Meno-Lite-0.1 \\
\midrule
IFEval, strict & \textbf{0.791} & 0.679 \\
GSM8K, flexible / strict & 0.719 / 0.187 & \textbf{0.759} / \textbf{0.619} \\
MMLU, 5-shot & \textbf{0.743} & 0.721 \\
MMLU, Russian & 0.649 & \textbf{0.653} \\
JSONSchemaBench (easy), valid JSON & 0.302 & \textbf{0.981} \\
Evidence exact match under PoT (DocSem) & 0.646 & \textbf{0.934} \\
Librusec history (Russian QA) & 0.781 & \textbf{0.906} \\
\bottomrule
\end{tabular}}
\caption{Cross-task comparison of the base and the adapted 7B model. Rigid output formats split the pair sharply in favor of the adapted model, while free-form instruction and English world knowledge favor the base; Russian narrative knowledge favors the adapted model.}
\label{tab:lineage}
\end{table}

First, the pair diverges most on rigid output formats, and the divergence is language-independent: the base model emits valid JSON only 30\% of the time on the easy JSONSchemaBench split (0.302 against 0.981), follows the GSM8K boxed-answer convention (``\#\#\#\#'') in 19\% of cases (0.187 against 0.619), and, once it writes programs, drops to 0.646 evidence exact match where the adapted model holds 0.934.
The citation and format discipline that PoT rewards inside DocSem is thus a transferable property of structured-output training, in English as much as in Russian.
Second, the language trade-off is real but narrower than the design hypothesis: the base keeps a small edge in English world knowledge (0.743 vs.\ 0.721 MMLU), the two are within noise on Russian MMLU (0.649 vs.\ 0.653), and the adapted model leads only in Russian narrative knowledge (0.906 vs.\ 0.781 on Librusec history), while the base wins free-form instruction (0.791 vs.\ 0.679).
Extraction is the one axis without a clean winner (NEREL-Bench \citep{bondarenko2026nerelbench}: the base leads entity recognition 0.473 vs.\ 0.437 F1, the adapted model relation extraction 0.248 vs.\ 0.192).
What the adaptation bought, then, is not ``more language'' in general but a specific, portable competence in structured output---exactly what a program-writing harness consumes.
Domain adaptation cost the model 0.022 of English world knowledge (0.743 to 0.721 MMLU) and bought 0.679 of JSON validity and 0.432 of strict-format compliance on GSM8K.
Structured-output competence is therefore not a side benefit but a precondition for harness-based scaling: PoT pays where that discipline exists (Meno-Lite: +0.282) and taxes where it does not (base Qwen2.5-7B: $-0.194$).
JSONSchemaBench validity (0.981 vs.\ 0.302) predicts the sign of the PoT effect better than parameter count does.

\section{What Happened on Test}
\label{sec:test}

\begin{table}[t]
\centering\footnotesize
\resizebox{\columnwidth}{!}{%
\begin{tabular}{lccc}
\toprule
Test attempt & Joint (\%) & Ans. (\%) & Evid. F1 \\
\midrule
1: Tesseract reading & 0.29 & 2.54 & 1.50 \\
2: VLM page transcription & \textbf{13.58} & \textbf{17.57} & \textbf{20.34} \\
3: + merge, mm escalation, repair & 13.41 & 17.34 & 20.11 \\
\bottomrule
\end{tabular}}
\caption{Portal truth for the three accepted attempts, in percent. The leaderboard selected attempt 2; rank 149 of 163.}
\label{tab:attempts}
\end{table}

\begin{table}[t]
\centering\footnotesize
\begin{tabular}{lccc}
\toprule
Diagnostic per attempt & 1 & 2 & 3 \\
\midrule
Singleton evidence, \% & 93.9 & 64.5 & 71.9 \\
Three or more blocks, \% & 3.8 & 32.5 & 25.3 \\
Evidence codes outside doc, \% & 8.7 & 3.5 & 1.7 \\
\bottomrule
\end{tabular}
\caption{Submission diagnostics. Gold evidence is always a single block; inflated sets mark a model that cannot locate the passage.}
\label{tab:diagnostics}
\end{table}

Table~\ref{tab:attempts} shows the portal truth; Table~\ref{tab:diagnostics} the diagnostics we computed afterwards.
The first attempt, built on Tesseract, was effectively a random draw (2.54\% answers).
Replacing the reader with page-wise vision transcription multiplied joint accuracy by five, and answers changed on 89.6\% of tasks between the two attempts: the difference came from re-reading the documents, with the reasoning stack unchanged.
The third attempt bundled everything that had passed our check-split gates: the three-vote merge with multimodal escalation, plus a deadline repair of the 38 pages the audit had flagged.
It altered 14.3\% of submission rows relative to attempt 2 and scored 0.17 percentage points of joint accuracy lower, a difference within sampling noise on 1{,}730 tasks; the informative part is that a 14\% turnover of predictions produced no gain to trade against.
The repair itself changed nine rows, all inside the 26 re-read documents, several of them degrading a nonzero answer to zero on the repaired text.

Format sanitation, the one axis that clearly improved across attempts (invalid evidence codes fell from 8.7\% to 1.7\% of tasks), did not move the metrics, which locates the remaining failure in semantics: wrong passage, wrong numbers, or both.
Evidence inflation points the same way: on a third of tasks the model returned three or more blocks where the gold set has exactly one.

The leaderboard is bimodal.
A smooth distribution would follow if reasoning architecture separated teams; instead the field splits into a majority that read the raster documents well enough to reason, a dense 22-team spike at 67.46\% joint accuracy, and a long tail that did not---a shape consistent with reading quality, not reasoning, having separated the field, though we cannot verify other teams' pipelines.
The top-ranked team reached 85\% with the same inputs, so the task was solvable and the failure sat in our reading stage.

\section{A Controlled Degradation Study}
\label{sec:degradation}

The post-mortem is observational; we can also run the experiment it implies.
We re-rendered the validation PDFs to match the measured physical properties of the test scans: 72-dpi grayscale JPEG (the test pages embed 69--78-dpi JPEG images), with a diagonal ``TESTING COPY'' stamp and a running header, both baked into the pixels as in the genuine article.
A median document in this degraded corpus shows a stopword share of 0.237, between the born-digital originals (0.349) and the test scans as read by the same OCR (0.153), so the treatment lands between the two regimes it connects.
The same pipeline, the same 72B generator, and the same sampling as the 0.825 validation row then run over the degraded rendering; the organizers' portal, which holds the validation labels, scores the result.

Table~\ref{tab:degradation} reports the outcome, and it splits cleanly by reader.
Tesseract collapses exactly as on the real test: answers fall from 0.825 to 0.244, and evidence falls to exactly zero, because OCR garbles the printed \texttt{b06:} prefixes into tokens like \texttt{0\%:} that no longer match the gold identifiers at all.
The vision-language reader recovers almost everything on our synthetic scans: 0.756 joint against 0.825 on born-digital input, with perfect evidence identifiers, because page-wise transcription restores the canonical prefixes it can read from context.
The genuine test told a different story for the same reader (17.6\% answers), and the distance between these two numbers is itself a finding: a clean re-render at test-like resolution reproduces the OCR failure mode but underestimates what the real scans did to vision transcription, whose damage on test came from dirtier scans, eight-page documents with four times the blocks, and identifier tokens that cannot be restored from convention.
The controlled half of the claim stands (rendering alone can zero out the primary metric through the identifier channel), and the uncontrolled half sharpens the lesson of Section~\ref{sec:data}: simulate the inputs, then verify the simulation against the real thing before trusting either.

\begin{table}[t]
\centering\footnotesize
\resizebox{\columnwidth}{!}{%
\begin{tabular}{lccc}
\toprule
Reading regime & Joint & Ans. & Evid. F1 \\
\midrule
val, born-digital (reference) & 0.825 & 0.825 & 1.000 \\
val, degraded raster + Tesseract & 0.000 & 0.244 & 0.000 \\
val, degraded raster + VLM pages & 0.756 & 0.756 & 1.000 \\
\midrule
test, Tesseract (attempt 1) & 0.29 & 2.54 & 1.50 \\
test, VLM pages (attempt 2) & 13.58 & 17.57 & 20.34 \\
\bottomrule
\end{tabular}}
\caption{The same tasks and model under three input renderings, scored by the validation portal (top, fractions) against the two real test attempts (bottom, percent). Synthetic degradation reproduces the OCR collapse; the residual gap to the real test rows measures what the simulation does not capture.}
\label{tab:degradation}
\end{table}

\section{Related Work}
\label{sec:related}

DocSem instantiates the GSM-SEM recipe \citep{singh2026gsmsem} over synthetic documents; our reasoning stack follows Program-of-Thoughts \citep{chen2023program} and self-consistency \citep{wang2023selfconsistency}, built on chain-of-thought prompting \citep{wei2022cot} and the program-aided line of work \citep{gao2023pal}, with structured output served by vLLM \citep{kwon2023efficient}.
Our system is a retrieval-augmented generation pipeline \citep{lewis2020rag}; graph-based retrieval descends from GraphRAG and its lightweight variants \citep{edge2024graphrag,guo2025lightrag,gutierrez2024hipporag}, evaluated for multi-hop settings by \citet{xiang2026graphragbench}; our finding is a boundary condition: on single-passage arithmetic over short documents, only the distilled top-$k$ of the graph earns its tokens.
Ensemble selection by a lightweight judge follows our SemEval-2026 system \citep{bondarenko-etal-2026-raguteam}; here the analogous device, majority voting over context variants, did not beat the best single vote.
Grounding over tabular and textual evidence connects to HybridQA and TaPas \citep{chen-etal-2020-hybridqa,herzig-etal-2020-tapas}; the DocSem twist is the exact-match block identifier, which punishes any segmentation drift.

\section{Conclusion: Lessons with Numbers}
\label{sec:conclusion}

First, audit the physical nature of evaluation inputs before the architecture.
We verified checksums, duplicates, and label recoverability, and skipped the pixel-level look that would have shown watermarked raster scans; the cost was the gap between 0.853 and 0.136 joint accuracy.
Second, application architecture is worth more than model scale, because the two kinds of competence a model can supply do not scale alike: routing world knowledge to external tools leaves the model only the linguistic residue, where a 27B model with the full harness matches a 72B model without it at roughly 2.7$\times$ fewer parameters, on one GPU and 0.14\,kgCO$_2$e against eight GPUs and 0.53\,kgCO$_2$e.
None of this transfers from born-digital validation to damaged test reading, which is why the first lesson comes first.
Third, the lever is model-dependent in a predictable way: PoT pays where format discipline exists and taxes where it does not, which is how a compact model is made harness-ready rather than merely small.
Fourth, deadline repairs treat symptoms.
The re-read of 38 damaged pages changed nine submission rows and slightly lowered the score, because the visible artifacts were a small sample of the systemic transcription damage.

\section*{Limitations}

This is a single-shared-task study; the reading-failure analysis rests on portal scores and submission diagnostics, not on human re-annotation of test documents.
Our vision reader was a 7B model transcribing pages independently; we did not evaluate stronger document-parsing VLMs, so the ceiling of our pipeline on raster inputs is unknown.
The 7B lineage comparison runs an English benchmark through a Russian-primary model, which is the weaker side of its declared domain; the comparison bounds the harness, not the model.
The external-judge probe referenced in the system description covers 60 test tasks only.
Validation-portal numbers rest on 217 tasks (95\% confidence intervals of four to five points at the observed accuracies); differences below one point between validation rows should be read accordingly.

\section*{Ethics Statement}

The system processes only the shared-task PDFs and outputs numeric answers with block identifiers; no personal data is processed. Model serving used institutional cluster resources.

\bibliography{references}

\appendix

\section{Run Registry}
\label{app:registry}
All 181-task check rows referenced in Table~\ref{tab:main}, with sampling parameters, appear in the internal run registry; the val rows of Table~\ref{tab:val} are portal-scored.
PoT demonstrations (34 verified programs) lift evidence exact match to 1.000 and cost 0.02--0.03 answer accuracy: a trade we declined for the final system.
A verification pass that asks the model to re-derive constants from cited blocks subtracts 0.013 joint by rejecting honest fractions.

\section{Retrieval Benchmark}
\label{app:retrieval}
Evidence recall on the check split: the bge-m3 pair reaches 0.348/0.646/0.983 at top-1/3/8 and 0.807 after reranking to top-3; the gte pair reaches 0.901/0.995/1.000 and 1.000.
A component ablation of the gte pair shows the dense vectors carrying the selection: BM42 alone reaches recall@3 of 0.066, because paraphrase queries are built to avoid the wording of the target passage, and the reranker lifts dense top-8 to a perfect top-3.
Track C (mixing vector and graph search contexts) won by 0.022 joint at $k{=}3$ on the compact line and was neutral at $k{=}5$; the graph-native track B lost 0.11 joint to vector retrieval on these short documents.

\section{Cost Accounting}
\label{app:cost}
Mini-graph extraction for the test submission (1{,}730 tasks, two extractor calls per retrieved passage) ran at about four tasks per minute on a ten-instance 7B fleet, roughly seven hours end to end; generation over the same tasks at $k{=}5$ took thirteen minutes per four-document shard on a single 27B server. The full-document baseline of Table~\ref{tab:main} skips the retrieval stack entirely, at the price of prompts that grow with document length.

Measured serving costs for the 181-task check runs of Table~\ref{tab:main} (wall clock on our infrastructure, A100 80GB; energy estimated at 400\,W TDP per GPU, PUE 1.1, grid intensity 0.35\,kgCO$_2$e/kWh):

\begin{table}[h]
\centering\footnotesize
\resizebox{\columnwidth}{!}{%
\begin{tabular}{lcccc}
\toprule
Configuration & GPUs & Wall & GPU-h & kgCO$_2$e \\
\midrule
72B direct, $k{=}1$ & 8 & 15 min & 2.0 & 0.31 \\
72B PoT, $k{=}3$ & 8 & 26 min & 3.5 & 0.53 \\
72B PoT, $k{=}5$ & 8 & 26 min & 3.5 & 0.53 \\
72B full doc, $k{=}3$ & 8 & 29 min & 3.9 & 0.60 \\
7B PoT (Meno-0.1), $k{=}5$ & 1 & \menokewall{} & \menokegpu{} & \menokeco{} \\
\bottomrule
\end{tabular}}
\caption{Serving cost of the check runs. The compact line reaches 0.64 joint on a single GPU in about twice the wall time the 72B line spends on eight of them.}
\label{tab:efficiency}
\end{table}

\section{Prompts and Ontology}
\label{app:prompts}
The generator instruction fixes five rules: reason only from displayed blocks; never mix numbers across passages; the program may only read displayed block contents; the answer is a canonical decimal string; evidence identifiers are copied character by character.
In the PoT mode the same instruction asks for a Python program whose last assignment holds the answer, executed by a sandboxed interpreter with a 2\,s budget; samples whose programs fail to execute are dropped from the vote.
The extractor receives 18 entity types (the NEREL numeric and object core \citep{loukachevitch2021nerel} plus QUANTITY and RATE) and 13 relation types (PRICE\_OF, INCOME, EXPENDITURE, AGE\_IS, POINT\_IN\_TIME, START\_TIME, END\_TIME, PART\_OF, LOCATED\_IN, TAKES\_PLACE\_IN, AGENT, OWNER\_OF, HAS\_QUANTITY), with MEASURED\_IN reserved for unit relations.

\section{Degradation Protocol}
\label{app:degradation}
Each validation page is rendered to 72-dpi grayscale, stamped, re-rendered 1:1, and stored as JPEG quality 55 in a one-image-per-page PDF, so the watermark and header live in the pixels and the text layer is empty, matching the genuine test scans (69--78~dpi JPEG).
The shadow snapshot keeps original file names, so the pipeline resolves tasks unchanged; only the parse differs.
Parsing runs in generic identifier mode, because OCR corrupts the printed \texttt{b01:} prefixes into tokens like \texttt{0\%:} or \texttt{`b12}; strict known-mode parsing then returns zero blocks. Unchanged paths also invalidate the path-to-hash index unless cleared.

\end{document}